# Character Segmentation in Asian Collector's Seal Imprints: An Attempt to Retrieval Based on Ancient Character Typeface

**Kangying Li [1*], Biligsaikhan Batjargal [2], Akira Maeda [3]**

[1] Graduate School of Information Science and Engineering, Ritsumeikan University, Japan

[2] Kinugasa Research Organization, Ritsumeikan University, Japan

[3] College of Information Science and Engineering, Ritsumeikan University, Japan

*Corresponding author: gr0319ss@ed.ritsumei.ac.jp

**Abstract**
Collector's seals provide important clues about the ownership of a book. They contain much information pertaining to the essential elements of ancient materials and also show the details of possession, its relation to the book, the identity of the collectors and their social status and wealth, amongst others. Asian collectors have typically used artistic ancient characters rather than modern ones to make their seals. In addition to the owner's name, several other words are used to express more profound meanings. A system that automatically recognizes these characters can help enthusiasts and professionals better understand the background information of these seals. However, there is a lack of training data and labelled images, as samples of some seals are scarce and most of them are degraded images. It is necessary to find new ways to make full use of such scarce data. While these data are available online, they do not contain information on the characters' position. The goal of this research is to provide retrieval tools assist in obtaining more information from Asian collector's seals imprints without consuming a lot of computational resources. In this paper, a character segmentation method is proposed to predict the candidate characters' area without any labelled training data that contain character coordinate information. A retrieval-based recognition system that focuses on a single character is also proposed to support seal retrieval and matching. The experimental results demonstrate that the proposed character segmentation method performs well on Asian collector's seals, with 92% of the test data being correctly segmented.
**keywords**
Asian seal imprint; Ancient document image processing; Character segmentation

# INTRODUCTION

The use of collector's seals in antique Asian books is a worthy topic to be discussed in detail. These seals express a sense of ownership, demonstrate inheritance, show identity, symbolizes social status and wealth, display aspiration and interest, identify edition, and express speech, amongst others, in different forms. For individual collector, in addition to their names, these seals also contain other words that depict their personal or family situations, including ancestral origin, residence, family background, family status, rank, and so on. Natsume Soseki, a famous Japanese writer, had many kinds of collector's seals. Foreign books, which he collected throughout his lifetime, were recorded with various collector's seals that he used in different periods. One can obtain vast information about his book-collecting habits by analyzing the seals. One of the most important functions of a collector's seal is to record where a book has been kept and the footsteps of its history of being handed over. For book-collecting institutions, the relocation process of books and their purchase histories can also be reflected in the contents of the seals. The names of many book-collecting institutions have been gradually changed, and through collector's seals, one can understand the books themselves, their collectors, and further background information on the collecting institutions. Through collector's seals in libraries, the collective experience and the source of inheritance of a book can be determined.

In Asian countries where kanji characters are used, ancient characters have typically been used to make collector's seals. As time passes, the shape of the characters might change, and multiple variations of a character might be created. There are also characters derived from a single or several kanji that look just like kanji characters. For instance, Vietnam's 'Chữ Nôm' is a character system that originated from kanji's original shape. Therefore, it is hard for non-professionals to understand all the contents of a collector's seal or sometimes even a single character in it.

When a single ancient character is recognized, typically by utilizing the data from scholars studying kanji characters, people can also gain a broader understanding of an ancient character's culture. Especially for individual collectors, exploring the information regarding every single character of their names is crucial to obtain a comprehensive understanding of their family affairs.

The objective of this study is to construct a retrieval-based ancient character recognition system that can match a user's query character, even if there is only one labelled typeface image, and update a new character category at any time instead of re-training a model.

**I RELATED WORK**
[Fujitsu R&D Center, 2016] proposed a seal retrieval technique that differs from related technologies in that it is specifically for Chinese antique document images. A color separation technique is used to separate the seal from the background image and then a two-level hierarchical matching method based on global feature matching is applied. [Su et al., 2019] introduced a seal imprint verification system featuring edge difference that uses a support vector machine (SVM) to classify the proposed feature. This system is aimed at the entire content of a seal, so the retrieval scope depends on the existing seal features in the database. [Sun et al., 2019] proposed a Chinese seal image character recognition method based on graph matching. In this method, a skeleton feature extracted from each character is used to construct the graph, graph matching is then applied to calculate the correspondence matrix, and the most similar reference character is selected. A single character database is used in this process, but the segmentation method has not been explained in detail.

Most seals are personal assets and tend to consist of meaningful, separate, and unique characters, so it makes sense to extract a single character from a seal and then analyze it individually. There have been several studies on character segmentation in antique documents. [Zahan et al., 2018] proposed a segmentation process for a printed Bangla script, and it has a good performance in segmenting characters with topologically connected structures; however, the target script is quite different from the target script in our research. [Nguyen et al., 2016] proposed a two-stage method to segment Japanese handwritten texts that utilizes vertical projection and stroke width transform for coarse segmentation and bridge finding and Voronoi diagrams for fine segmentation. While this method performs very well for the segmentation of Japanese handwritten characters, seal characters usually have irregular positions or distributions, as well as notable differences in character size, which may have a negative impact on the segmentation result when using the proposed method. [Ren et al., 2011] have put forward several character segmentation methods for both circular and elliptical Chinese seal imprints. One of these methods fits the contour of a seal image into a geometrical shape (e.g., a circle or an ellipse), then uses mathematical transformation to convert it into a rectangular region, and finally obtains a single character according to the calculation of the horizontal distribution. [Liu et al., 2007] also introduced a character segmentation method for Chinese seals that focuses on two types of circular seals. However, since the collector's seal data in the present study have irregular edges and irregular character distribution, a concrete means to segment and extract a single character from various seals with different forms of edges is needed.

In this work, we propose a text extraction method for the image data of the collector's seals. The proposed method is intended for use in future research involving automatic text feature generation to explore the background information of antique Asian books from external databases using extracted text contents in character recognition tasks. We developed a character segmentation method to deal with irregular character distributions in seals with different shapes and perform character segmentation using mean-shift clustering. Our retrieval method is proposed by using the features extracted from ancient font typefaces and obtained ranking results through calculations of similar characters.

## II METHODOLOGY

We describe our approach in the following sections. Since we only extract features from typeface images and store them in a database, we perform pre-processing on user-provided seal images, which is described in Section 2.1. In Section 2.2, we explain the character segmentation method, and in Section 2.3, we go over the process of extracting essential features from images. Section 2.4 describes the extraction of features from the font typeface. The feature matching and ranking calculation are respectively introduced in Sections 2.5 and 2.6.

### 2.1 Data pre-processing
As shown in Fig.1, in classical Asian materials, seals often overlap with handwritten words.

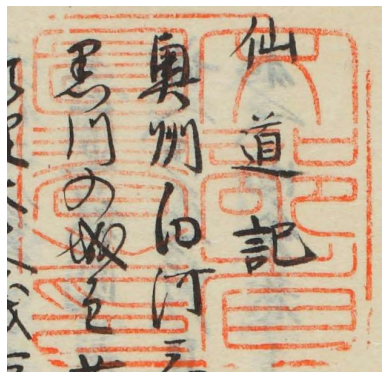

Figure 1. Example of a seal overlapping with handwritten words.

Therefore, extracting the seal from the image is an important task. We use k-means clustering [Hartigan et al., 1979] to cluster image color information. As shown in Fig.2, we project the information of three RGB channels of an image into a three-dimensional space.

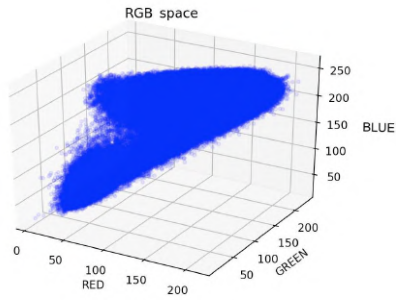

Figure 2. Three-dimensional representation of color information.

Hence, the Euclidean distance is used to represent the color relationship between two pixels, which is defined by

$$D_{rgb} = \sqrt{(R_1 - R_2)^2 + (G_1 - G_2)^2 + (B_1 - B_2)^2}, \qquad (1)$$

where $R_1, R_2, G_1, G_2, B_1, B_2$ represent RGB values for pixels 1 and 2, respectively, and $D_{rgb}$ is used to cluster pixels with similar colors. According to the principle of the k-means algorithm, we regard this task as extracting K groups of pixels with similar colors from images. Our system automatically extracts areas with more red components. As shown in Fig.3, pixel group 2 is extracted as the analysis target.

| Clustering results | Pixel group 1 | Pixel group 2 | Pixel group 3 |
|---|---|---|---|
| (RGB space, K=3 plot) | (handwritten text) | (seal pattern) | (outline text) |

Figure 3. Results of k-means clustering.

A lot of data pertaining to signatures and seals have recently become available due to the increase of digital archiving. With this data increase in mind, as shown in Fig.4, the proposed method provides users with an interactive interface that enables them to choose any part of a seal they want to retrieve rather than automatically selecting the part without handwritten text as the result.

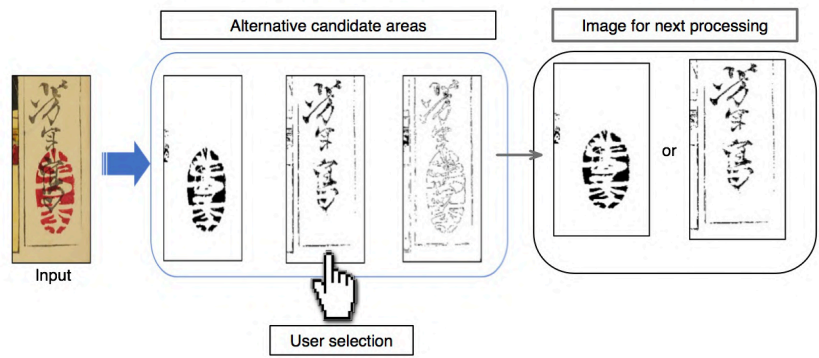

Figure 4. Selection of candidate results.

In order to improve the calculation speed, we normalize the selected input image size to the range [(0×0), (200×200)] using min-max normalization, as

$$Size_{output} = 200 \times \left( \frac{Size_{original_{input}} - Max(Size_{original})}{Max(Size_{original}) - Min(Size_{original})} \right), \qquad (2)$$

where $Size_{original_{input}}$ represent the length or width of the original image and $Size_{output}$ shows the corresponding results. $Size_{original}$ represents the set {length, width} of the original image.

## 2.2 Character segmentation
The main shapes of the seal image we focus on are shown in Fig.5.

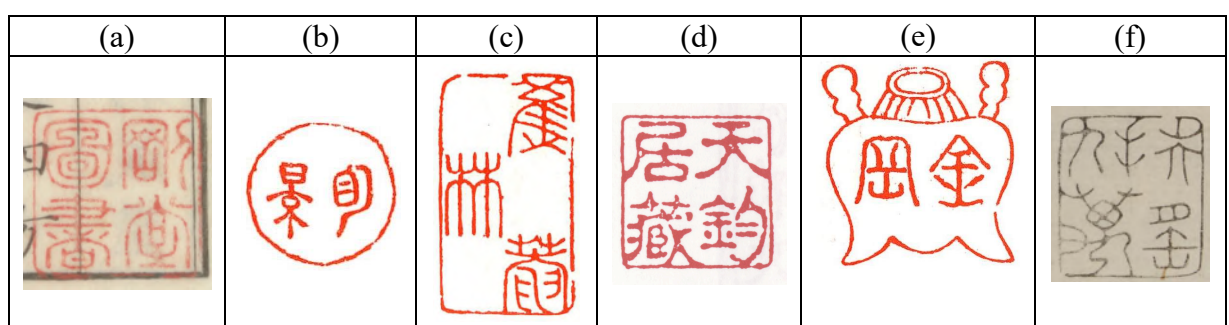

Figure 5. Seal shapes.

These shapes include (c) rectangles, (a)(d)(f) squares, (b) circles, and (e) seals with irregular edges. Our proposed method is suitable for seal imprints in which the boundary of each character can be identified by non-experts. It was impossible for us to recognize the segmentation ground truth of some of the seals because they featured irregular characters and images, sometimes even mixed together, and only experts would be able to identify the segmentation ground truth. To ensure that we can check the effectiveness of the proposed method ourselves, in this study we examined only seals that have an easily recognizable layout. Examples of some seals that are not included in this paper are provided in Fig.6.

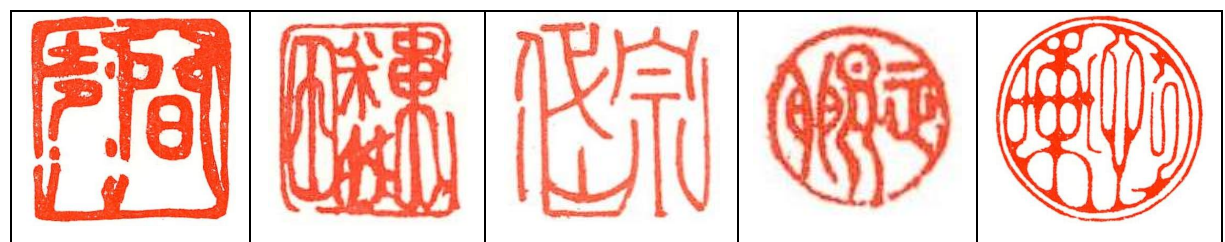

Figure 6. Seals outside the scope of this research.

For the determination of a single character, we use clustering to segment each character. Because kanji characters are independent and balanced in structure, we regard every character as a module, and each module has its own centroid. By considering these centroids, we can use clustering to extract character fields. Since we already know the coordinates of each pixel of the seal areas, the density information of each pixel can be obtained by kernel density estimation. Values are calculated by

$$\hat{f}_{bandwidth(X,Y)}(x,y) = \frac{1}{n}\sum_{i=1}^{n}\frac{1}{2\pi bandwidth(X,Y)} exp\left(-\frac{((x-X_i)^2+(y-Y_i)^2)}{2\pi bandwidth(X,Y)^2}\right), \quad (3)$$

where $X$ and $Y$ are the vectors $X = \{x_0, x_1, ... x_n\}$, $Y = \{y_0, y_1, ... y_n\}$ extracted from the foreground pixel $set\{(x_0, y_0), (x_1, y_1) ... (x_n, y_n),\},$ $n$ is the number of elements in vector $X$ or $Y$, $X_i$ is the $i_{th}$ element in vector $X$, and $Y_i$ is $i_{th}$ element in vector $Y$. $bandwidth(X,Y)$ refers to the optimal bandwidth values.

We use mean-shift clustering [Comaniciu et al., 1999] to cluster the pixels of an image. The clustering results can be optimized by adjusting the bandwidth to consider different variables. Fig.7 shows the results of clustering under different bandwidth settings when the input is not a normalized image.

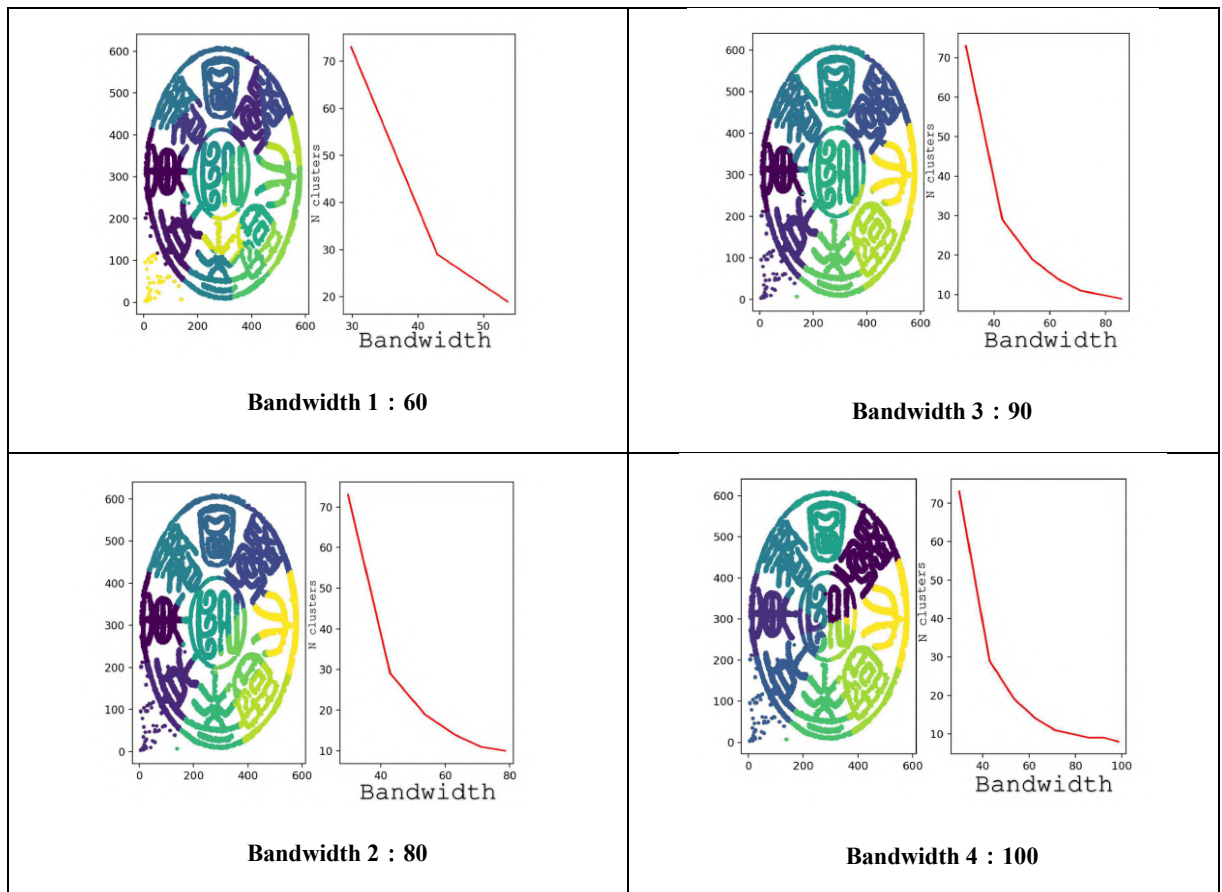

Figure 7. Visualization of clustering results under different bandwidths.

In each unit, the graph on the left side is the clustering result of each pixel of the seal, and each color represents a different cluster. For the graph on the right side, the X-axis is the value of the bandwidth and the Y-axis is the number of clusters.

The results we need are selected as follows. When the change rate of the total number of clusters becomes steady, for example, when the bandwidth is about 90, we treat each cluster in the result as a candidate result of the character segmentation. As shown in Fig.8, we calculate the bandwidth interval, and the bandwidth value is obtained equidistantly in the interval. These bandwidth values are then used to obtain the segmentation candidates.

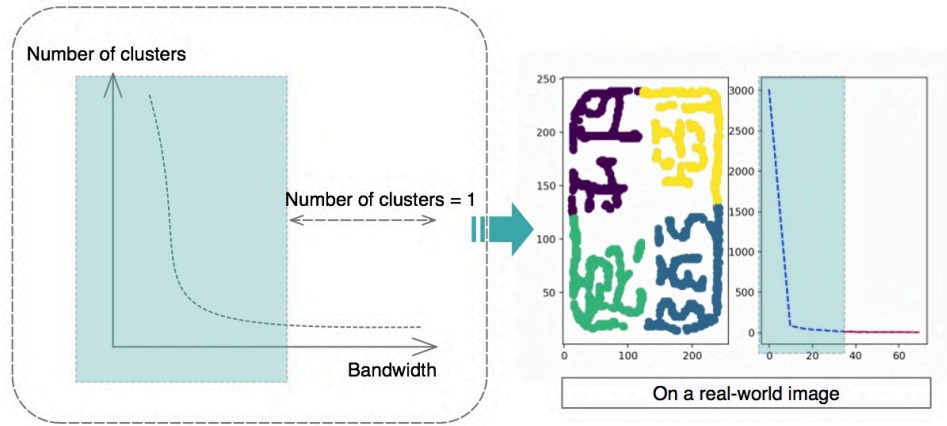

Figure 8. Visualization of clustering results under different bandwidths.

The algorithm is shown in Algorithm 1.

| Algorithm 1 |
|---|
| **Input**: |
|     **F**: coordinate set $\{(x_0, y_0), (x_1, y_1) \ldots (x_n, y_n),\}$ from foreground pixel obtained by K-means clustering |
|     **R**: Randomly generated sorted set $\{r_1, r_2, r_3, r_4 \ldots r_i,\}$, $r_i \in [0.0,1.0]$ for bandwidth estimation, default i=100 |
| **Output**: Set of character location hypotheses **U** |
| 1:    Compute the length of **F**, $len_F$=len(**F**) |
| 2:    Initialize set of estimate_bandwidth **Bandwidth**={} |
| 3:    **For each** $r_1 \in \mathbf{R}\{r_1, r_2, r_3, r_4 \ldots r_i,\}$ **do**: |
|         Set the number of neighbors=$r_i * len_F$, compute the nearest neighbor for each coordinate in **F** |
|         For each coordinate in **F**, compute farthest neighbor and distance between them, get a set of farthest distance |
|         $\mathbf{Distance_{farthest}} = \{dis_{farthest}(x_0, x_0), dis_{farthest}(x_1, x_1) \ldots dis_{farthest}(x_n, x_n)\}$ |
|         Compute the average of $\mathbf{Distance_{farthest}}$ get $Average_{(r_i, farthest\ neighbour)} = \frac{\sum_0^i dis_{farthest}(x_0, x_0)}{len_F}$ |
|         Add $Average_{(r_i, farthest\ neighbour)}$ into **Bandwidth** |
| 4:    For each $\{b_1, b_2, b_3, b_4 \ldots b_i,\}$ in **Bandwidth**, get the result of number of clusters |
|     $\{Nclusters_{b1}, Nclusters_{b2}, Nclusters_{b3}, \ldots Nclusters_{bi}: Nclusters_{bi} \in f_{meanshift_{clustering}}(b_i)\}$ |
| 5:    Fit a polynomial $Q(b)$ with set $\{(b_1, Nclusters_{b1}), (b_2, Nclusters_{b2})\ldots(b_i, Nclusters_{bi})\}$ |
| 6:    Get the second derivative $Q''(b) = \frac{d^2 Nclusters_{bi}}{db_i^2}$ of $Q(b_n)$ |
| 7:    Initialize set of Descent_bandwidth **Descentband**={} |
| 8:    **For each** $\{b_1, b_2, b_3, b_4 \ldots b_i,\}$ in **Bandwidth**, |
|         IF $Q''(b_i) < 0$ |
|         Add $b_i$ into **Descentband** |
| 9:    Get sorted **Descentband** = $\{b_{\mathbf{Descent\_1}}, b_{\mathbf{Descent\_2}}, \ldots b_{\mathbf{Descent\_n}}\}$ |
| 10:   Set an interval $Interval_{descentband}$, $default = 5$ |
| 11:   Group **Descentband** by $Interval_{descentband}$, get set |
|     $\mathbf{G_{interval}} = \{Group_{descentband1}, \ldots Group_{descentbandn}, (\{b_{\mathbf{Descent}_{n-\ Interval_{descentband}}}, b_{\mathbf{t}_{n-\ Interval_{descentband}}+1}, \ldots \} \in Group_{descentbandn})\}$ |
| 12:   Compute the standard deviation of each group in $\mathbf{G_{interval}}$, get $STD_{group} = \{std_{group1}, std_{group2} \ldots std_{groupn}\}$ |
| 13:   Compute the minimum value of $STD_{group}$, get the group **Candidates** from $\mathbf{G_{interval}}$ corresponding to the minimum value |
| 14:   Initialize set of character location hypotheses **U** |
| 15:   **For each** $\{b_{candidate1}, b_{candidate2}, b_{candidate3}, \ldots b_{candidaten}\}$ in **Candidates**, |
|         Compute the result of clustering $f_{meanshift_{clustering}}(b_{candidaten_i})$, $i \in n$ |
|         Classify the **F** use the output labels of clustering, add the set $Character_{candidate\_i} =$ |
|         $\{(x_{label_{i\_0}}, y_{label_{i\_0}}), (x_{label_{i\_1}}, y_{label_{i\_1}}) \ldots (x_{label_{i\_n}}, y_{label_{i\_n}}) : (x_{label_{i_n}}, y_{label_{i_n}}) \in \mathbf{F}\}$, with same label $i$ into **U** |

As show in Fig.9, the segmentation results under adjacent bandwidth may have a large area of overlap.

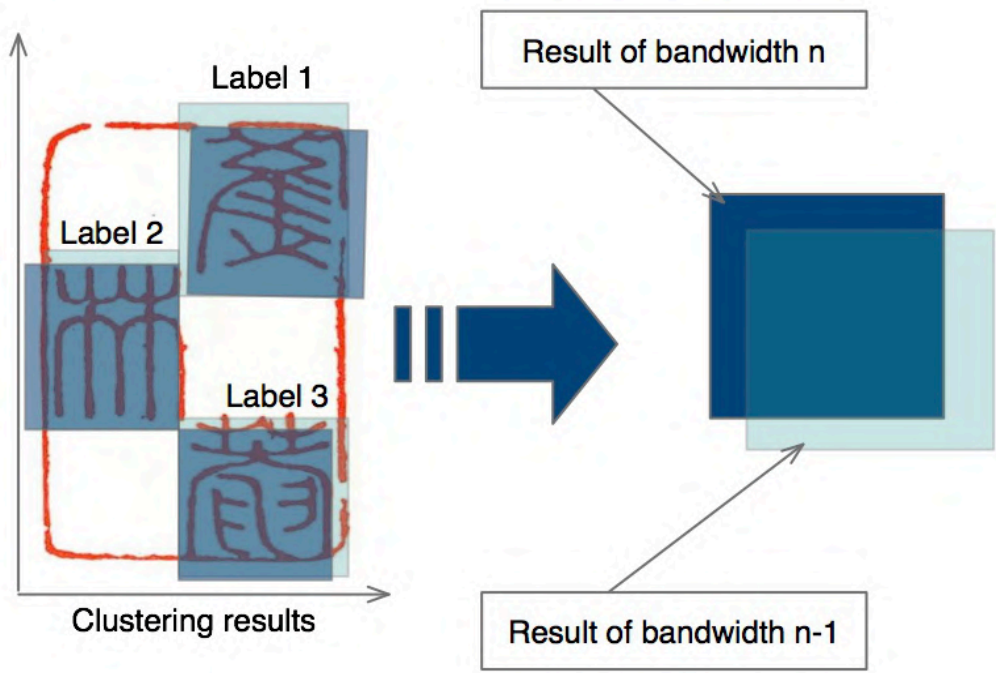

Figure 9. Clustering results under adjacent bandwidth.

We only keep the larger of the two candidate areas that overlap more than 90% of the area of any one of them. The algorithm implementation is available on GitHub [Li et al., 2019] as a reference.

**2.3 Extracting CNN features from images**

The typeface images extracted from the font file [Shirakawa Font project, 2016] are shown in Fig.10.

| *Modern commonly used characters* | '人' | '文' | '科' | '学' |
|---|---|---|---|---|
| *"Shirakawa font" Typeface* | ㇵ | 穴 | 𬳵 | 學 |

Figure 10. Images converted from a font file.

First, we normalize the typeface images. Next, we crop the typeface images in accordance with the maximum and minimum coordinates of the black pixels, and then standardize them to the size of 225 × 225. As there are many variations of ancient characters, even a slight change of the structure will affect the extraction of geometric features. We use the pre-trained model to extract the deep features (convolutional neural network (CNN) features) of fonts, with the aim of subtracting minor changes in a character's structure, to help move characters of the same category closer to each other in the feature space.

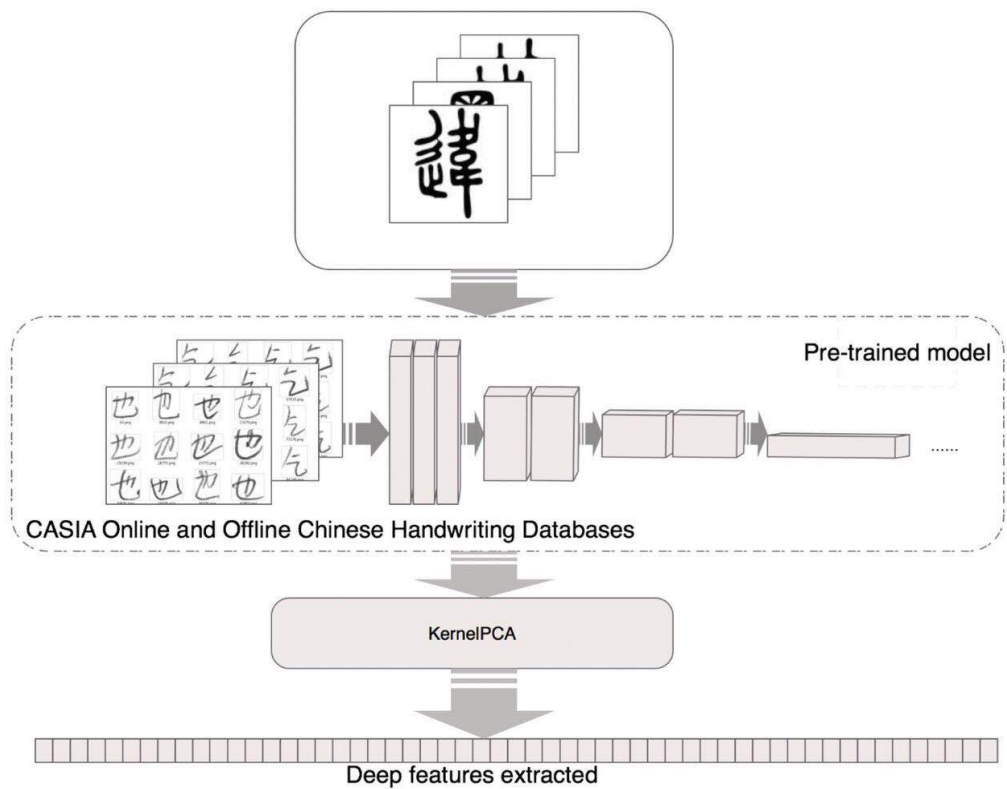

Figure 11. Extraction of CNN features.

As shown in Fig.11, due to the limited content in the ancient character dataset, in addition to the pre-trained model, which was trained by ImageNet, we also utilize CASIA Online and Offline Chinese Handwriting Databases [Liu et al., 2011], in which the characters have some shape features in common with ancient characters, as the training data to train the model using VGG16 [Simonyan et al., 2014]. The visual expression of the feature map in the max pooling layer of the pre-trained model is shown in Fig.12. We use kernelPCA [Mika et al., 1999] to reduce the dimensions of the output from the middle layer.

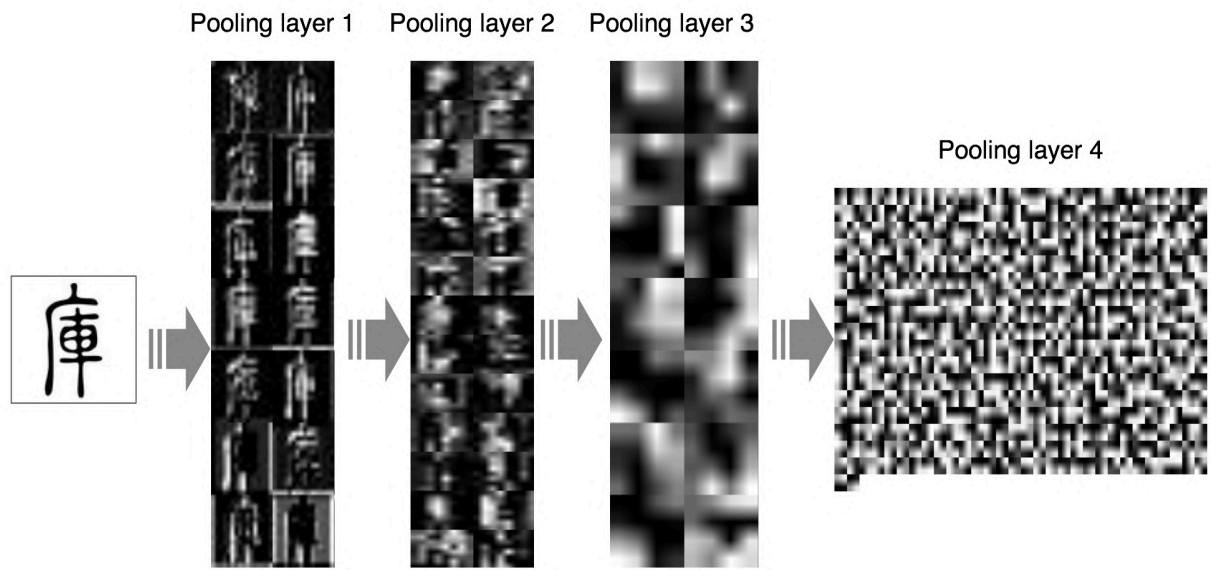

Figure 12. Visual expression of feature map in max pooling layer.

## 2.4 Extracting geometric features from images

To capture the details of a character's structure, we extract the geometric features of these characters.

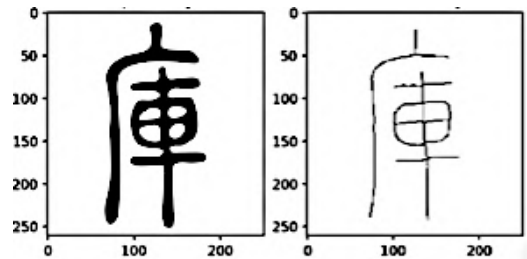

Figure 13. Skeleton map of a character.

We obtain a character's skeleton map by using the method of Zhang et al. [Zhang et al., 1984], as shown in Fig.13. In contrast to a general thinning method, Zhang's method ignores the existence of the stroke width and can obtain a skeleton map without noises, so each stroke can be represented by a unique corresponding continuous single pixel. Then we use the Harris corner [Harris et al., 1988] to obtain the coordinates of the intersection of each stroke. The coordinate points of the skeleton map and of the stroke intersections are then stored to database as one of the representations of geometric features.

## 2.5 Image matching using multiple features

We use the following method for calculating similarity to match the typeface image and the query image. The matching process is shown in Fig.14. The same feature extraction method is applied to the segmented user query image and the typeface image to extract the corresponding features (including CNN and geometric features).

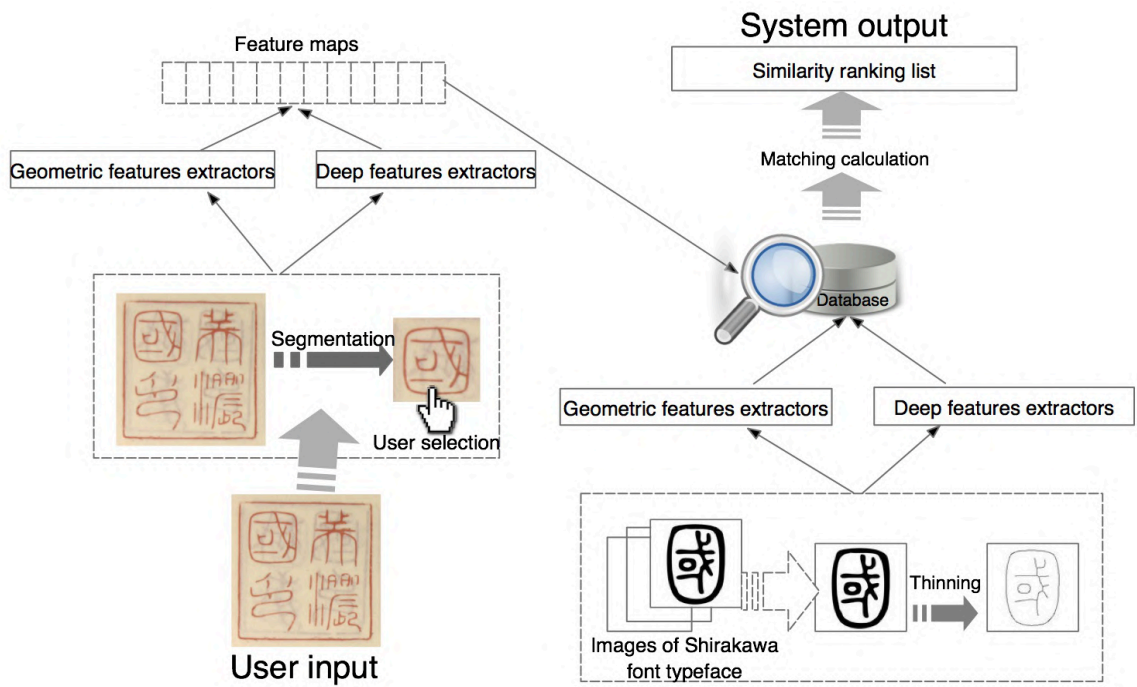

Figure 14. Matching process.

## 2.6 Ranking calculation

Since the features are different, and the value ranges are totally different, similarity scores of different features cannot be directly compared. We therefore convert the results of each similarity calculation into scores, and each feature is then given a different weight corresponding to its score. The results of distance are normalized to values in the range [0, 1] by using min-max normalization, and the function is defined by

$$S_{distance\_i} = \frac{distance_i - Min(\{distance_1, distance_2 ... distance_n\})}{Max(\{distance_1 ... distance_n\}) - Min(\{distance_1 ... distance_n\})},  \quad (3)$$

where $i$=0, 1...., $N$, $N$ is the number of images in the database, $S_{distance\_i}$ is the similarity score between image $i$ and the query image, and $distance_n$ is the distance between image $i$ and the query image calculated by using different distance formulas. The multi-feature similarity score is calculated by

$$S_{total} = \frac{W_{cf} S_{cnnFeature} + W_{gf} S_{geometricFeature}}{W_{cf} + W_{gf}}, \quad (4)$$

where $S_{total}$ is the total score of similarity, $S_{geometricFeature}$ is the sum of similarities of geometric features calculated using distance formulas, and $W_{cf}$ and $W_{gf}$ are the weights of the similarity score of CNN feature $S_{cnnFeature}$ and geometric feature $S_{geometricFeature}$. Finally, we use the total score of similarity to predict the input image's category.

## III EXPERIMENTS AND RESULTS

In this section, we describe our experiments. Section 3.1 introduces the database used in the experiments, and the results of image segmentation and image retrieval are presented in Sections 3.2 and 3.3.

## 3.1 Datasets

We selected the test data from the Collectors' Seal Database [National Institute of Japanese Literature, available from 2011] for our experiments. This database contains 36,537 images of collector's seals, including only pictorial seals and seals carved in various types of calligraphy. The main color of the seals is red. Because the original data do not include the coordinate information of a single character, we counted the frequently appearing characters in the database and marked their position information as a small-scale experimental object.

## 3.2 Segmentation results

We used several different types of seals to test our proposed segmentation algorithm. The results are shown in Fig.15. Our method achieved good results for processing images with irregular character distribution. However, as indicated by Result 3, larger characters were segmented into the candidate areas of other characters, so further research on dealing with images with different character sizes and close character spacing is required.

| 1: Regular seal | Result 1 | 2: Characters with irregular glyph | Result 2 |
|---|---|---|---|
| 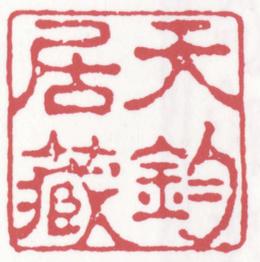 | 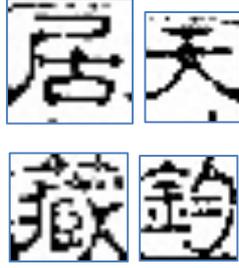 | 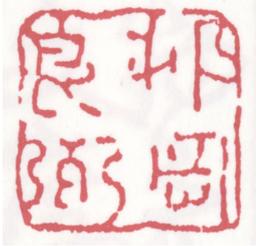 | 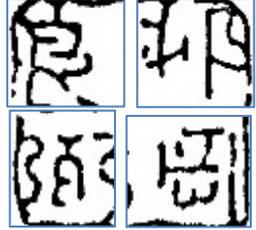 |
| 3: Irregular character distribution | Result 3 | 4: With noisy background and overlaps with handwritten words | Result 4 |
| 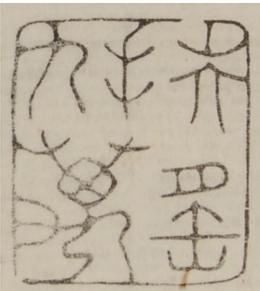 | 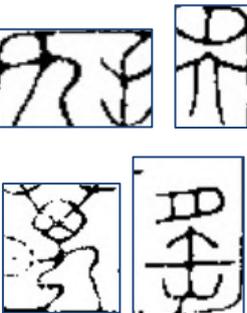 | 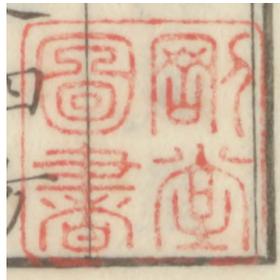 | 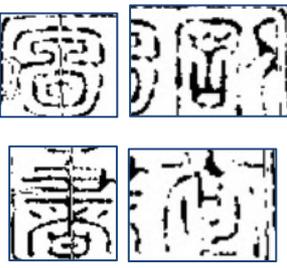 |

Figure 15. Segmentation results in square seal images.

The segmentation results for seals of different shapes are shown in Fig.16.

| 1: Rectangle | Result 1 | 2: Circle | Result 2 |
|---|---|---|---|
| 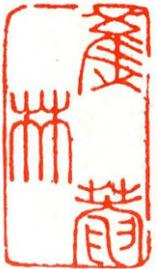 | 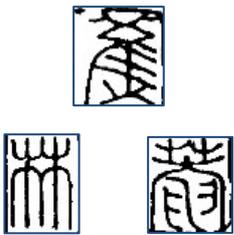 | 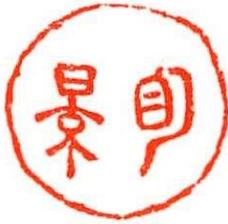 | 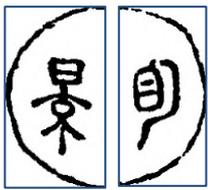 |
| 3: Irregular shape (a) | Result 3 | 4: Irregular shape (b) | Result 4 |
| 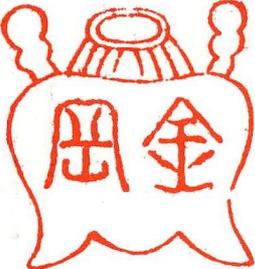 | 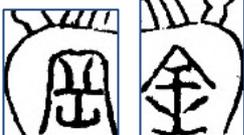 | 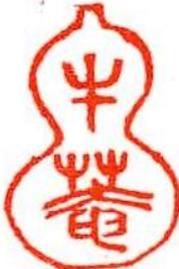 | 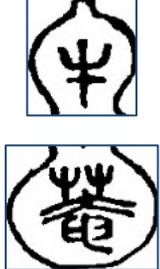 |

Figure 16. Segmentation results for seal imprint with different shapes.

These results demonstrate that our proposed method can deal with the seal imprints of different shapes.

**3.2 Retrieval results**

Because for each category of a character, we can only one converted typeface image, it is unfavorable to select the best features and method for distance calculation to classify characters with different shape. For the selection of features and distance calculation methods, we performed experiments on a standard dataset and selected Japanese Hiragana data from the Omniglot dataset [Lake et al., 2016], in which the character shapes are different from Chinese characters.

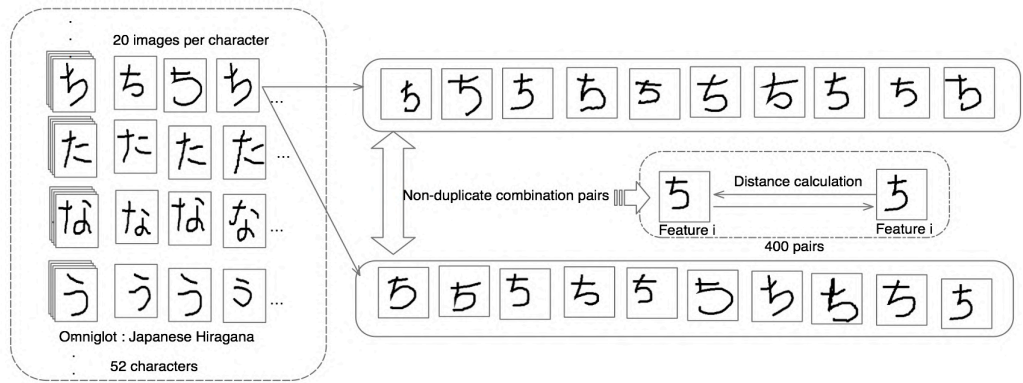

Figure 17. Experiment on feature and distance calculation method selection.

As shown in Fig.17, the experiments on feature selection were done by calculating the average similarity from the same character. We define the average similarity of the same character $q$ as $Sim_{ave\_q}$. There are 20 images in one character in the Omniglot dataset. We define $P_{character\_q}$ as a set of all the generated non-duplicate combination pairs from 20 images, and $Sim_{ave\_q}$ can be obtained by calculating the average similarity of $P_{character\_q}$. Hence, we evaluate the different features by

$$S_{average_{similarity}} = \frac{\sum_{q}^{L_{characterSet}} f_{feature\_i}(Sim_{ave\_q})}{L_{characterSet}}, \qquad (5)$$

where $S_{average_{similarity}}$ is the average similarity score for each feature, $f_{feature_i}(Sim_{ave\_q})$ is the average similarity of character $q$ in $feature_{\_i}$, and $L_{characterSet}$ is the total number of character categories.

*3.2.1 Feature selection*

We need to choose the best image performance under geometric features and the best layer and pretrained model to extract deep features. The original image and the skeleton map (introduced in Section 1.3), shown in Fig.18, were used in this experiment.

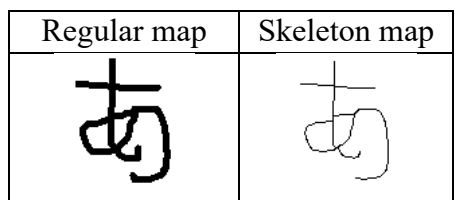

Figure 18. Regular map and skeleton map used in matching experiments.

We first conducted an experiment on geometric features. The histogram of oriented gradients (HOG) [Dalal et al., 2005] is one of the most efficient descriptors to extract precise and invariant local and global features [Mikolajczyk et al., 2003] [Lim et al., 2010]. It enables the difference between similar characters to be distinguished by strokes, and the intersection of strokes is also an important feature. We used the Harris Corner Detector [Harris et al., 1988] to extract the difference between strokes.

Fig.19 shows the $Sim_{ave\_q}$ of each character, where the horizontal axis is the index of the character and the vertical axis is the average similarity of each character.

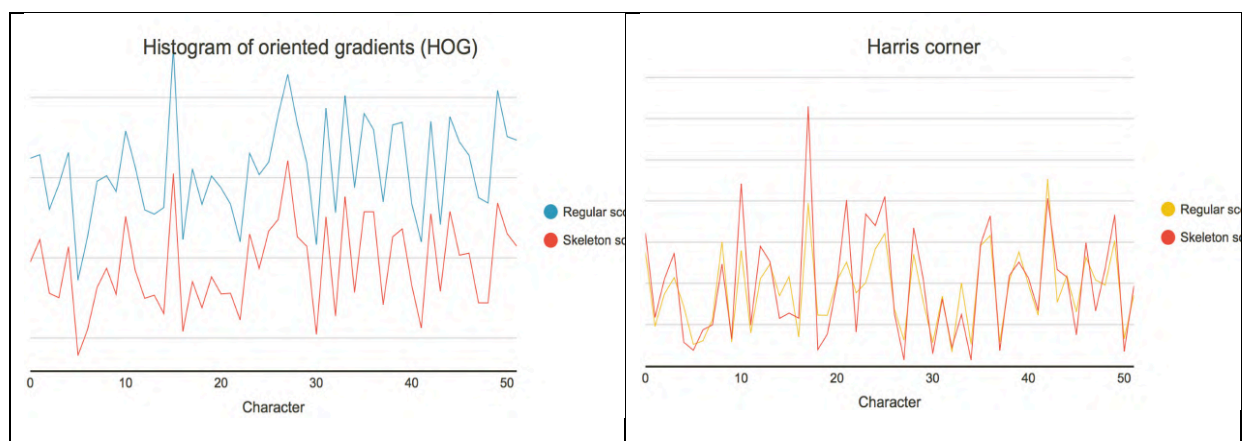

Figure 19. Experiment on feature and distance calculation method selection.

The results show that we can get better results by using the regular image in the HOG feature and the skeleton map to obtain the Harris corner feature.

In order to select CNN deep features, we utilized the VGG19 pre-trained model trained by ImageNet and the VGG16 pre-trained model trained by the handwritten Chinese characters dataset [Liu et al., 2011] as feature extractors. Features were extracted using the intermediate layer from 'pool3' to 'fc2' of these pre-trained models. Tables 1 and 2 show the $S_{average_{similarity}}$ in deep features extracted from the model trained by ImageNet and by the handwritten Chinese characters dataset, respectively.

|  | Regular map | Skeleton map |
| --- | --- | --- |
| **VGG19_fc1** | 0.79040205 | 0.82432610 |
| **VGG19_fc2** | 0.82494724 | 0.84423065 |
| **VGG19_pool3** | 0.58491635 | 0.59660584 |
| **VGG19_pool4** | 0.40395800 | 0.43494022 |
| **VGG19_pool5** | 0.53847855 | 0.58189917 |

Table 1. Evaluation of features extracted from pre-trained model by ImageNet.

|  | Regular map | Skeleton map |
| --- | --- | --- |
| **VGG16_fc1** | 0.97992086 | 0.99724620 |
| **VGG16_fc2** | 0.97555065 | 0.99678916 |
| **VGG16_pool3** | 0.99224720 | 0.99905205 |
| **VGG16_pool4** | 0.99565570 | 0.99944544 |
| **VGG16_pool5** | 0.97863317 | 0.99706230 |

Table 2. Evaluation of features extracted from pre-trained model by handwritten Chinese characters dataset.

The results show that the features extracted from the skeleton map had the highest similarity between characters.

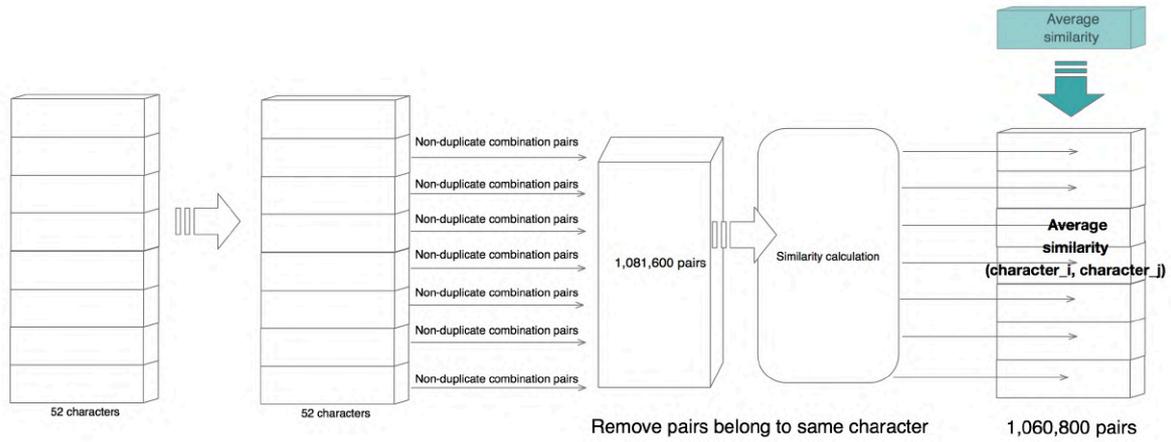

Figure 20. Calculating the distinction score of each character.

We also need to determine whether a certain feature can be distinguished from other characters. In Fig.21, we introduce the calculation of distinction score $Score_{distinction}$. Non-duplicate combination pairs were generated from all images from Hiragana character data. We calculated the average similarity between pairs where neither element of the pair belonged to the same character. Finally, we calculated the average results to represent $Score_{distinction}$. The evaluation of a feature $i$ is done by

$$Score_{feature\_i} = \left| S_{average_{similarity}} - Score_{distinction} \right|. \qquad (6)$$

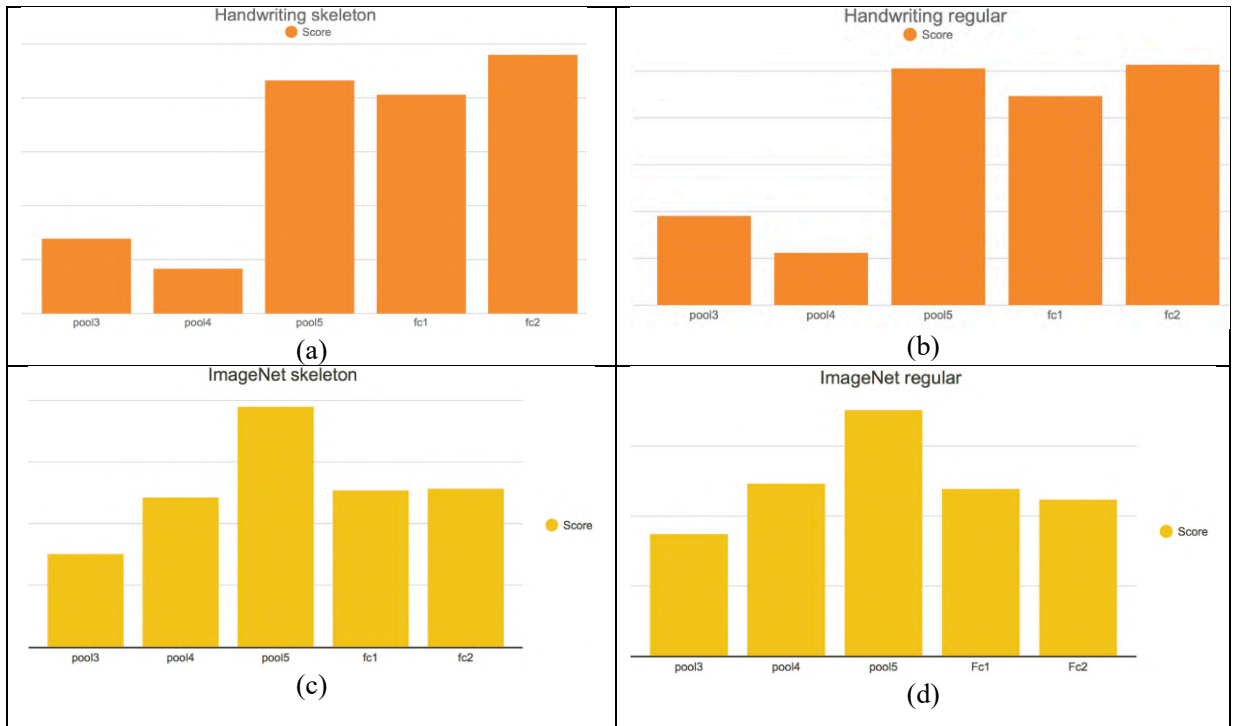

Figure 21. $Score_{feature\_i}$ in different situations. (a) and (b) show the distinction score of regular image and skeleton map in ImageNet pre-trained model. (c) and (d) show the distinction score in pre-trained model by CASIA Online and Offline Chinese Handwriting Databases.

The results in Fig.21 show the global image feature, which was extracted by the pool 5 layer that had the best performance.

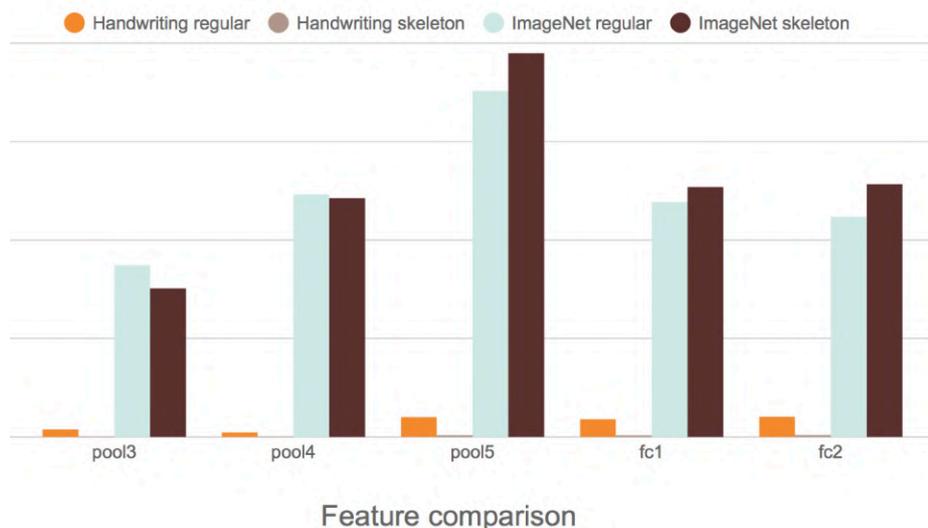

Figure 22. $Score_{feature\_i}$ in different situations.

Fig.22 shows a comparison of the $Score_{feature\_i}$ results in different situations. We can see that the features extracted from the ImageNet pre-trained model with skeleton maps had the best discrimination ability.

In Section 3.2.3, we will discuss the seal imprint retrieval results by using the features shown in Table 3.

| Feature | Input image |
|---|---|
| **Harris Corner Detector** | Regular map |
| **Histogram of oriented gradients** | Skeleton map |
| **VGG19_pool5(ImageNet)** | Skeleton map |

Table 3. Features used in seal imprint retrieval.

*3.2.3 Retrieval using seal imprint*

Here, we show the results of the ten characters with the highest frequency in the printed text, which were calculated by Mean Reciprocal Rank (7)

$$MRR = \frac{1}{|Q|} \sum_{i=1}^{|Q|} \frac{1}{rank_i}, \quad (7)$$

where $Q$ is the total number of images retrieved, $i$ is the image number, and rank is the ranking order. For each character, we tested with 20 images.

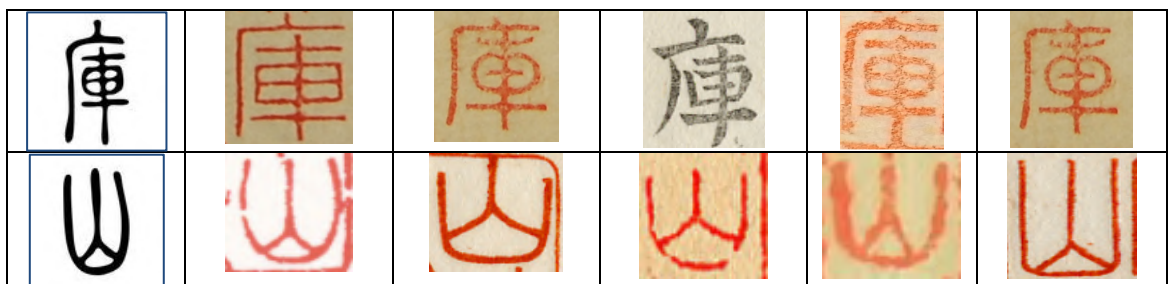

Figure 22. Part of the data used in retrieval experiments.

The results are shown in Fig.23.

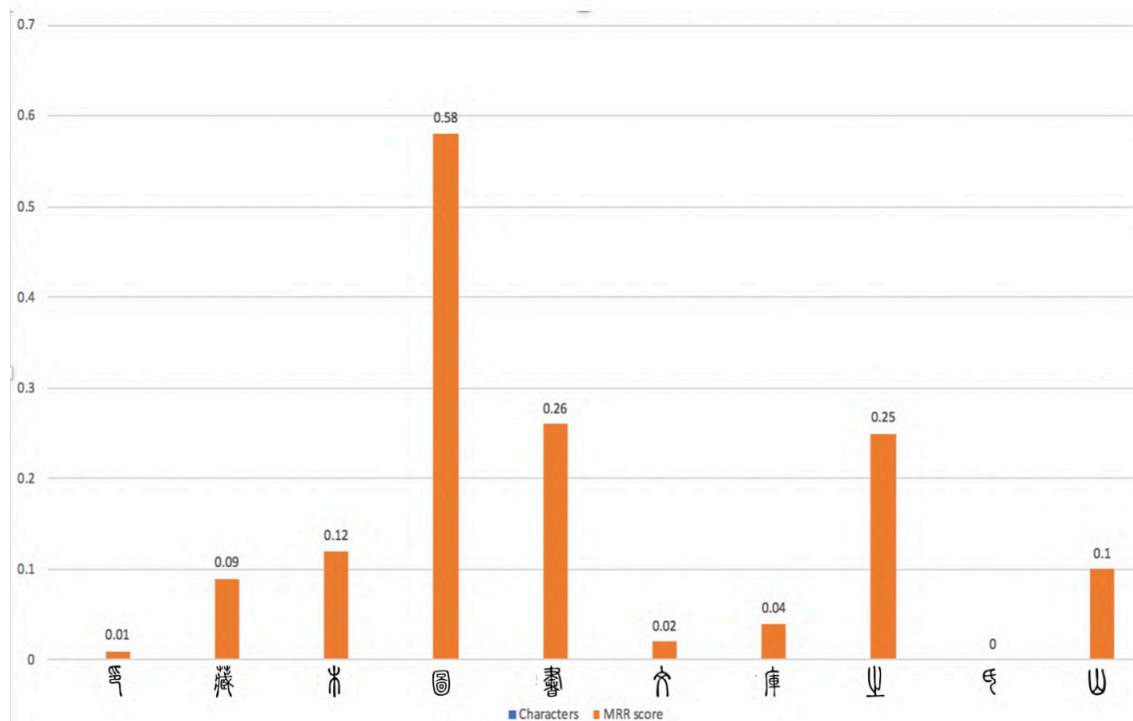

Figure 23. Evaluation on seal imprints.

There are 2,586 images in the Shirakawa font dataset, so our retrieval scope includes all 2,586 images in different categories. Most of the test data had a performance in the top-50 ranking results. Characters with less deformation were in the top-5 ranking results. However, as shown in Fig.24, the results of the similarity calculation were negatively affected when the query was significantly different from our standard typeface.

| Standard typeface | Query | | | | |
|---|---|---|---|---|---|
| 氏 | 氏 | 氏 | 氏 | 氏 | 氏 |

Figure 23. Examples of poor results.

## IV CONCLUSION

In this work, we used two clustering algorithms to pre-process seal images and obtained good results. In contrast to training a neural network, a clustering algorithm extracts part of the required information from an image without consuming a lot of computational resources. We used a combination of deep features and geometric features to retrieve ancient kanji characters through the calculation of similarities. This reduces the time takes to re-train a model when adding a new category of characters and also enables flexible usage of the intermediate output of neural networks. We can determine which seal belongs to which famous person by using the recognition results, which enables us to learn more about this person's habits from his or her collection information. Exploring this further will be the next target of our research. We will also investigate how to make better use of a single typeface image.